\documentclass[journal]{IEEEtran}  
\usepackage[utf8]{inputenc}

\usepackage{amsmath} 
\usepackage{amsfonts} 
\usepackage{amssymb}
\usepackage{subfig}
\usepackage{tikz}
\usepackage{float}
\usepackage{caption}
\usepackage{hyperref}
\usepackage{cleveref}
\usepackage{array}
\usepackage{textcomp}
\usepackage{stfloats}
\usepackage{url}
\usepackage{verbatim}
\usepackage{graphicx}
\usepackage{cite}
\usepackage{booktabs}

\usepackage{textgreek}
\usepackage{multirow}
\usepackage{arydshln}
\usepackage{algorithmic, algorithm} 

\begin{document}

\title{HTR-JAND: Handwritten Text Recognition with Joint Attention Network and Knowledge Distillation}

\author{Mohammed Hamdan, Abderrahmane~Rahiche,~\IEEEmembership{Member,~IEEE,} 
        Mohamed~Cheriet,~\IEEEmembership{Senior Member,~IEEE,}%
\thanks{Authors are with the Synchromedia laboratory, \'{E}cole de Technologie Sup\'{e}rieure (\'ETS), University of Quebec, Montreal, Canada.}
\thanks{Manuscript received October XX, 2024; revised XX XX, 2025.}}

\markboth{This paper is currently under review at IEEE Transactions}
{Hamdan \MakeLowercase{\textit{et al.}}: HTR-JAND: Handwritten Text Recognition with Joint Attention Network and Knowledge Distillation}


\maketitle


\begin{abstract}
The digitization and accurate recognition of handwritten historical documents remain crucial for preserving cultural heritage and making historical archives accessible to researchers and the public. Despite significant advances in deep learning, current Handwritten Text Recognition (HTR) systems struggle with the inherent complexity of historical documents, including diverse writing styles, degraded text quality, and computational efficiency requirements across multiple languages and time periods.
This paper introduces HTR-JAND (HTR-JAND: Handwritten Text Recognition with Joint Attention Network and Knowledge Distillation), an efficient HTR framework that combines advanced feature extraction with knowledge distillation. Our architecture incorporates three key components: (1) a CNN architecture integrating FullGatedConv2d layers with Squeeze-and-Excitation blocks for adaptive feature extraction, (2) a Combined Attention mechanism fusing Multi-Head Self-Attention with Proxima Attention for robust sequence modeling, and (3) a Knowledge Distillation framework enabling efficient model compression while preserving accuracy through curriculum-based training.
The HTR-JAND framework implements a multi-stage training approach combining curriculum learning, synthetic data generation, and multi-task learning for cross-dataset knowledge transfer. We enhance recognition accuracy through context-aware T5 post-processing, particularly effective for historical documents. Comprehensive evaluations demonstrate HTR-JAND's effectiveness, achieving state-of-the-art Character Error Rates (CER) of 1.23\%, 1.02\%, and 2.02\% on IAM, RIMES, and Bentham datasets respectively. Our Student model achieves a 48\% parameter reduction (0.75M versus 1.5M parameters) while maintaining competitive performance through efficient knowledge transfer. Source code and pre-trained models are available at \href{https://github.com/DocumentRecognitionModels/HTR-JAND}{Github}.
\end{abstract}


\section{Introduction}
\IEEEPARstart{H}{andwritten} text recognition in historical documents represents a cornerstone of digital humanities and cultural heritage preservation. The ability to accurately convert handwritten documents into machine-readable text is essential for making centuries of historical records, manuscripts, and cultural artifacts accessible to researchers, historians, and the public. This task presents significant challenges due to writing style variability, document degradation, and diverse linguistic content across multiple time periods \cite{fischer2010ground, bluche2017gated}. Figure~\ref{fig:sample_images} illustrates these challenges through representative samples from different historical periods and writing styles, highlighting the complexity of developing robust recognition systems.

Traditional approaches based on segmentation methods \cite{graves2008novel, bianne2011dynamic} and complex processing pipelines \cite{dutta2018improving, plotz2009markov} struggle with capturing the nuanced relationships between handwriting styles and textual content. These methods often require extensive preprocessing and manual intervention, limiting their applicability in large-scale digitization projects. Current deep learning methods, while promising, face three fundamental limitations: inconsistent generalization across writing styles and historical periods \cite{fischer2011transcription, chowdhury2018efficient}, difficulties in handling long text sequences \cite{michael2019evaluating, puigcerver2017}, and computational requirements that restrict practical deployment \cite{tassopoulou2021enhancing, yousef2020accurate}. While attention mechanisms have improved sequence modeling capabilities \cite{kang2020pay, wick2021transformer}, existing approaches continue to struggle with balancing recognition accuracy and computational efficiency.

These challenges are further compounded by the lack of robust mechanisms for handling historical character variations and archaic writing styles \cite{fischer2010ground}. Combined with the computational demands of processing handwritten text recognition \cite{Hamdan2023Sep, hamdan2022refocus}, these limitations highlight the need for an integrated approach that addresses both accuracy and efficiency requirements.

To address these challenges, we present HTR-JAND (HTR-JAND: Handwritten Text Recognition with Joint Attention Network and Knowledge Distillation), an end-to-end framework that combines efficient feature extraction with knowledge transfer capabilities. Our approach includes several key components:

\begin{figure}[h]
    \centering
    \fbox{\includegraphics[width=0.9\columnwidth]{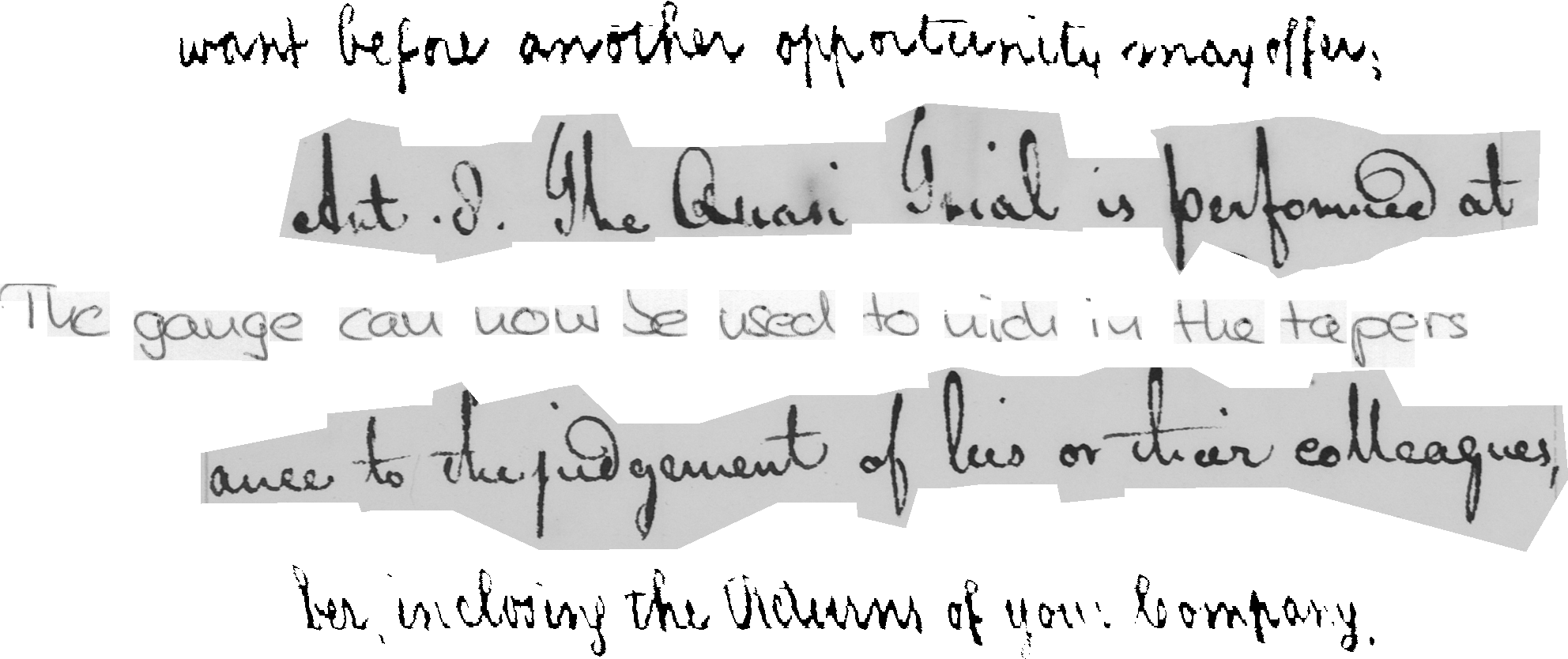}}
    \caption{Sample images from different datasets, demonstrating the range of challenges including writing style variability, non-standard character shapes, and contextual dependencies.}
    \label{fig:sample_images}
\end{figure}

\begin{itemize}
\item A comprehensive preprocessing pipeline that combines character set unification across datasets with adaptive oversampling, achieving balanced representation while maintaining a unified vocabulary of 103 characters across diverse historical periods and writing styles.

\item A CNN architecture combining FullGatedConv2d layers with Squeeze-and-Excitation blocks for adaptive feature extraction, inspired by recent advances in visual recognition \cite{deSousaNeto}.

\item A Combined Attention mechanism that integrates Multi-Head Self-Attention with Proxima Attention, building upon successful approaches in sequence modeling \cite{chowdhury2018efficient, kang2020pay}.

\item A knowledge distillation framework enabling compact model deployment while maintaining performance, extending techniques for model compression \cite{you2017learning}.

\item Training strategies incorporating curriculum learning with synthetic data generation \cite{Wigington2017Nov}, ensemble learning, and multi-task learning.

\item Context-aware post-processing using a fine-tuned T5 model to improve recognition accuracy in historical texts.

\item Comprehensive evaluations on standard benchmarks described in subsection \ref{subsec:preprocess} demonstrate HTR-JAND's effectiveness. The framework achieves state-of-the-art Character Error Rates of 1.23\%, 1.02\%, and 2.02\% on IAM, RIMES, and Bentham datasets respectively, while maintaining practical efficiency through significant parameter reduction.

\end{itemize}

The paper is structured as follows: Section \ref{sec:related} reviews recent HTR developments; Section \ref{sec:methodology} details the model architecture and loss function design; Section \ref{sec:advanced_training} describes training strategies; Section \ref{sec:results_discussion} presents experimental results; and Section \ref{sec:conclusion_future_work} concludes the paper with findings and future directions.

\section{Related Work}
\label{sec:related}

Handwritten Text Recognition (HTR) has seen significant advancements with deep learning techniques, and this section offers an overview of key developments by showcasing architectural innovations and identifying gaps our work addresses; Table \ref{tab:related_studies} summarizes key studies focusing on architectural innovations, attention mechanisms, and performance on benchmark datasets, with notes defining abbreviations: GC (Gated Convolution), SE (Squeeze-and-Excitation Blocks), CA (Combined Attention), KD (Knowledge Distillation), CL (Curriculum Learning), AR (Aspect Ratio Preservation), and PP (Post-processing).

\begin{table}[h]
\centering
\parbox{0.45\textwidth}{\caption{Overview of studies showing different architectural components implemented}\label{tab:related_studies}}
\begin{tabular}{lccccccc}
\hline
\textbf{Study} & \textbf{GC} & \textbf{SE} & \textbf{CA} & \textbf{KD} & \textbf{CL} & \textbf{AR} & \textbf{PP} \\ 
\hline
Graves et al. \cite{graves2006connectionist} & \checkmark & & & & & & \\ 
\hline
Puigcerver \cite{puigcerver2017} &  & & & & & & \\ 
\hline
Bluche \cite{bluche2017gated} & \checkmark & & & & & & \\ 
\hline
Chowdhury et al. \cite{chowdhury2018efficient} &  & & \checkmark & & & & \\ 
\hline
Kang et al. \cite{kang2020pay} &  & & \checkmark & & & & \\ 
\hline
Wigington et al. \cite{Wigington2017Nov} & & & & & \checkmark & & \\ 
\hline
Hamdan et al. \cite{Hamdan2023Sep} &  & \checkmark & & & & & \\ 
\hline
Flor et al. \cite{deSousaNeto} & \checkmark & \checkmark & & & & & \checkmark \\ 
\hline
Retsinas et al. \cite{retsinas2024best} &  & & & & & \checkmark & \\ 
\hline
(HTR-JAND) this Work & \checkmark & \checkmark & \checkmark & \checkmark & \checkmark & \checkmark & \checkmark \\ 
\hline
\end{tabular}
\end{table}

\subsection{Architectural Evolution in HTR}

The foundation of modern HTR systems was laid by traditional Hidden Markov Models (HMM) \cite{bianne2011dynamic,espana2010improving,plotz2009markov}, which provided probabilistic frameworks for sequence modeling but struggled with long-range dependencies and required careful feature engineering. This was followed by Graves et al. \cite{graves2006connectionist} introducing Connectionist Temporal Classification (CTC), enabling end-to-end training on unsegmented sequence data. This work, utilizing Bidirectional Long Short-Term Memory (BLSTM) networks, marked a significant departure from traditional HMM approaches by allowing the model to learn feature representations directly from raw input data.

Subsequent research focused on sequence-to-sequence modeling \cite{sueiras2018offline,zhang2019sequence,aberdam2021sequence}, which treated HTR as a translation problem from image to text, and integrating Convolutional Neural Networks (CNNs) with Recurrent Neural Networks (RNNs). These approaches enabled more flexible handling of variable-length inputs and outputs while capturing both spatial and temporal dependencies. Puigcerver \cite{puigcerver2017} demonstrated the effectiveness of CNN-LSTM architectures combined with CTC loss, achieving competitive results on standard benchmarks. This approach set a new baseline for HTR systems, balancing feature extraction capabilities with sequential modeling.

Further architectural innovations emerged to address specific challenges in HTR. Bluche \cite{bluche2017gated} introduced gated convolutional layers to better handle long text sequences, while Dutta et al. \cite{dutta2018improving} employed Spatial Transformer Networks to address geometric distortions in handwritten text. However, the challenge of handwriting variability, especially in historical documents, continues to impact model generalization \cite{fischer2010ground}.

\subsection{Attention Mechanisms and Advanced Techniques}

Attention mechanisms \cite{bahdanau2014neural,yang2016stacked,gregor2015draw,chen2018pixelsnail} have become increasingly prominent in HTR, allowing models to focus on relevant parts of the input during recognition. These mechanisms dynamically weight different regions of the input based on their relevance to the current prediction, enabling more precise character recognition and better handling of complex layouts. Self-attention approaches \cite{cheng2016long,parikh2016decomposable} further enhanced this capability by calculating responses at particular sequence locations by attending to the entire sequence, effectively capturing global dependencies without the need for recurrent connections. Chowdhury et al. \cite{chowdhury2018efficient} and Kang et al. \cite{kang2020pay} demonstrated the effectiveness of attention in end-to-end neural models and Transformer architectures, respectively, though capturing long-range dependencies in very long text sequences remains challenging \cite{kang2020pay}.

Recent works have explored more sophisticated techniques to improve HTR performance. Data augmentation strategies \cite{poznanski2016cnn,chammas2018handwriting} have proven effective for handling limited data scenarios, incorporating techniques such as elastic distortions, affine transformations, and synthetic data generation to improve model robustness. These methods have been particularly valuable for historical document recognition where training data is scarce. Hamdan et al. \cite{Hamdan2023Sep} incorporated Squeeze-and-Excitation (SE) blocks to enhance feature representation, while Flor et al. \cite{deSousaNeto} combined gated convolutions with SE blocks and introduced post-processing techniques. Retsinas et al. \cite{retsinas2024best} focused on preserving aspect ratios of input images, addressing the issue of information loss during preprocessing.

\subsection{Efficiency and Learning Strategies}

As HTR models grew in complexity, research began to focus on improving efficiency and generalization. Wigington et al. \cite{Wigington2017Nov} highlighted the importance of data augmentation and curriculum learning strategies. Knowledge distillation, as demonstrated by You et al. \cite{you2017learning}, emerged as an effective technique for transferring knowledge from large teacher models to smaller, more efficient student models. However, balancing computational efficiency with recognition accuracy, particularly for deployment in resource-constrained environments, remains an ongoing concern \cite{vaswani2017attention, Wick2021Sep}, and addressing data scarcity for historical or less common languages continues to challenge the field \cite{fischer2011transcription}.

As shown in Table \ref{tab:related_studies}, existing approaches have typically focused on individual components or techniques in isolation. Our work uniquely integrates multiple state-of-the-art techniques while introducing new elements. We combine gated convolutions with SE blocks for enhanced feature extraction, integrate a novel combined attention mechanism for improved handling of long-range dependencies, and implement knowledge distillation alongside curriculum learning strategies for better efficiency and generalization. The preservation of aspect ratios and advanced post-processing techniques further enhance our model's ability to handle diverse handwriting styles and complex documents. This comprehensive approach represents a significant step toward more robust and efficient HTR systems, addressing multiple challenges concurrently rather than in isolation.

\section{Methodology}
\label{sec:methodology}

Our approach to Handwritten Text Recognition (HTR) introduces several key innovations to address the challenges of diverse writing styles, historical documents, and computational efficiency. This section details our methodological contributions, emphasizing the novel aspects of our architecture and training strategy.

\subsection{Data Preprocessing and Augmentation}
\label{subsec:preprocess}

\begin{figure*}[h]
    \centering
    \includegraphics[width=\textwidth]{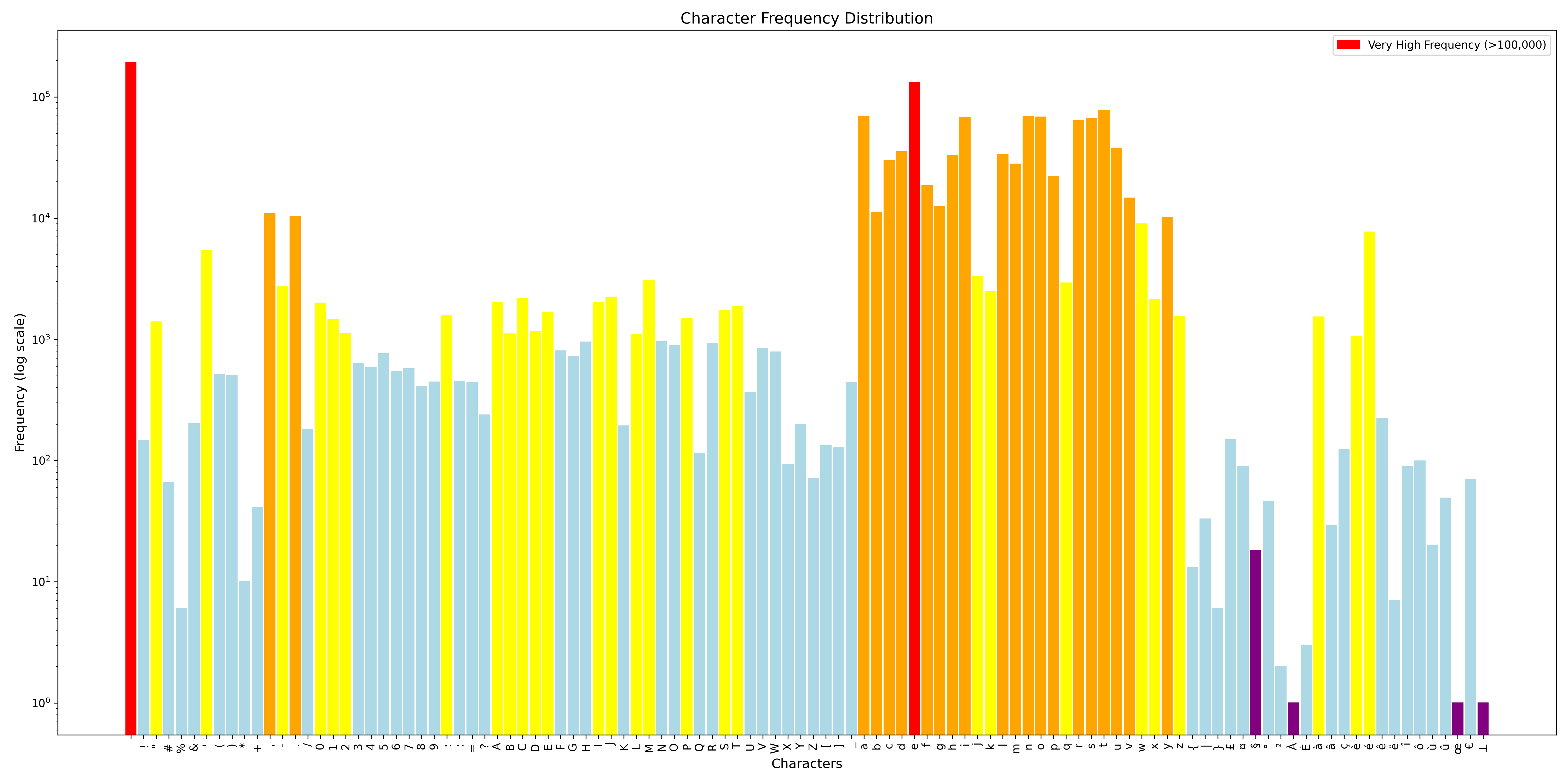}
    \caption{Distribution of character frequencies across the combined datasets. Note the removal of infrequent characters such as `\S', `\`A', and `\`oe'.}
    \label{fig:char_distribution}
\end{figure*}

To facilitate knowledge distillation and standardize training across multiple datasets including including IAM \cite{marti2002iam}, RIMES \cite{grosicki2009results}, Bentham \cite{causer2012building}, Saint Gall \cite{fischer2010ground}, and Washington \cite{fischer2011transcription}, our preprocessing approach begins with character set unification. The process removes infrequent characters that would not significantly impact classifier performance, resulting in a unified set of 103 unique characters across all datasets. As shown in Figure~\ref{fig:char_distribution}, character frequencies exhibit considerable variation, with some characters appearing frequently (lowercase letters and spaces) while others occur rarely.

Table~\ref{tab:dataset_stats} presents key statistics for each dataset after preprocessing, including a buffer of 2 added to the maximum sequence length to accommodate variations during inference.

\begin{table}[h]
    \centering
    \caption{Dataset statistics after preprocessing}
    \label{tab:dataset_stats}
    \begin{tabular}{lccccc}
        \hline
        Dataset & Train & Valid & Test &  Vocab & Max Len +2  Buffer \\
        \hline
        IAM & 6,161 & 900 & 1,861 & 79 & 93  \\
        RIMES & 10,193 & 1,133 & 778 & 100 & 110  \\
        Washington & 325 & 168 & 163 & 68 & 61 \\
        Bentham & 9,195 & 1,415 & 860 & 94 & 103  \\
        Saint Gall & 468 & 235 & 707 & 48 & 74  \\
        \hline
        Combined & 26,342 & 3,851 & 4,369 & 103 & 110 \\
        \hline
    \end{tabular}
\end{table}

Our preprocessing pipeline addresses three key challenges: handwriting style variability, limited labeled data availability, and temporal coherence preservation. Each input image $\boldsymbol{I}$ undergoes normalization to a standard size of $68 \times 864$ pixels:

\begin{equation}
\label{eq:normalize}
\boldsymbol{I}'_{x,y} = 2 \cdot \frac{\boldsymbol{I}_{x,y} - \min(\boldsymbol{I})}{\max(\boldsymbol{I}) - \min(\boldsymbol{I})} - 1.
\end{equation}

The complete preprocessing workflow follows Algorithm~\ref{alg:preprocess}:

Data augmentation applies transformations $\boldsymbol{T} = \{t_1, \ldots, t_n\}$ to each image:

\begin{equation}
\label{eq:augment}
\boldsymbol{I}_{\text{aug}} = t_n(\ldots t_2(t_1(\boldsymbol{I}))).
\end{equation}

The synthetic data generation process is defined in Algorithm~\ref{alg:synth}:

The pipeline incorporates three key strategies for training stability: curriculum-based synthetic ratio adjustment, performance-based adaptive synthetic data integration with a 10\% initial ratio, and enhanced augmentation techniques.

For class balancing, we implement adaptive oversampling:

\begin{equation}
\label{eq:class_balance}
w_c = \max(1, \frac{\bar{f}}{\epsilon + f_c}),
\end{equation}

where $w_c$ represents the sampling weight for character $c$, $f_c$ is its frequency, $\bar{f}$ denotes mean character frequency, and $\epsilon$ prevents division by zero.

This comprehensive approach ensures effective preprocessing across our diverse dataset while maintaining consistency and stability in the training process.
\subsection{The Proposed Model}
\label{sec:model_architecture}

\begin{algorithm}[t]
\caption{Preprocessing Pipeline with Synthetic Data (PPS)}
\label{alg:preprocess}
\begin{algorithmic}[1]
\renewcommand{\algorithmicrequire}{\textbf{Input:}}
\renewcommand{\algorithmicensure}{\textbf{Output:}}
\REQUIRE $D$, $C$, $F$, $r$, $\alpha$
\ENSURE $D'$
\STATE $\mathbf{D}_n \gets \text{Normalize}(D)$ \COMMENT{Eq. \ref{eq:normalize}}
\STATE $\mathbf{D}_a \gets \text{Augment}(\mathbf{D}_n)$ \COMMENT{Apply transforms}
\STATE $\mathbf{D}_s \gets \text{GenerateSynthetic}(C, F, r)$ \COMMENT{Algo \ref{alg:synth}}
\STATE $\mathbf{D}_t \gets \text{Tokenize}(\mathbf{D}_a \cup \mathbf{D}_s, C)$
\STATE $\mathbf{D}' \gets \text{BalanceClasses}(\mathbf{D}_t, \alpha)$
\RETURN $\mathbf{D}'$
\end{algorithmic}
\end{algorithm}

\begin{algorithm}[t]
\caption{Synthetic Data Generation (SDG)}
\label{alg:synth}
\begin{algorithmic}[1]
\renewcommand{\algorithmicrequire}{\textbf{Input:}}
\renewcommand{\algorithmicensure}{\textbf{Output:}}
\REQUIRE $C$, $F$, $r$, $D$
\ENSURE $\mathbf{D}_s$
\STATE $n \gets |D| \cdot r / (1-r)$
\FOR{$i = 1$ to $n$}
    \STATE $t \gets \text{RandomText}(C)$
    \STATE $f \gets \text{RandomChoice}(F)$
    \STATE $\boldsymbol{I} \gets \text{RenderText}(t, f)$
    \STATE $\boldsymbol{I}_{\text{aug}} \gets \text{Augment}(\boldsymbol{I})$
    \STATE $\mathbf{D}_s \gets \mathbf{D}_s \cup \{(\boldsymbol{I}_{\text{aug}}, t)\}$
\ENDFOR
\RETURN $\mathbf{D}_s$
\end{algorithmic}
\end{algorithm}

The proposed HTR architecture addresses handwritten text recognition challenges through a hierarchical structure combining convolutional neural networks, recurrent layers, and attention mechanisms. As shown in Figure \ref{fig:archicture}, the architecture processes input text line images through an encoder-decoder pipeline, employing a Teacher-Student framework as described in the next subsection \ref{subsec:knowledge_distillation} to balance recognition accuracy with computational efficiency.

\subsubsection{Architecture Overview}
\label{subsec:arch_overview}

\begin{figure*}
\centering 
\fbox{
\includegraphics[width=\textwidth]{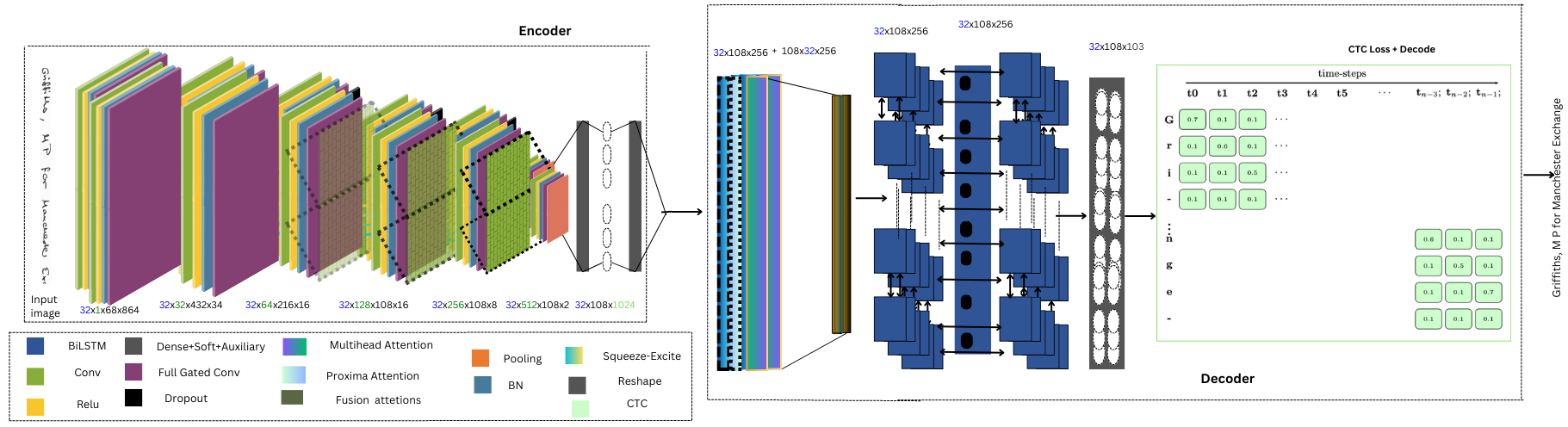} 
}
\caption[HTR Model Architecture]{Proposed HTR Model Architecture: Data flow through CNN feature extraction, LSTM sequence modeling, and Combined Attention mechanisms. Additionally, CTC Matrix for "Griffiths, M P for Manchester Exchange" showing probabilities for first "Gri-" and last "-nge" ('-' represents blank symbol for CTC alignment).}\label{fig:archicture}

\end{figure*}

The model employs a Teacher-Student framework where the Teacher model provides a comprehensive architecture that is later distilled into a more efficient Student model. The Teacher model integrates five key components, as illustrated in Figure \ref{fig:archicture}: CNN blocks with Squeeze-and-Excitation (SE) modules, FullGatedConv2d layers for adaptive feature extraction, bidirectional LSTM layers for sequence modeling, Multi-Head Self-Attention combined with Proxima Attention, and CTC-based decoding with auxiliary classification.

The model processes grayscale input images of size $68 \times 864$ through progressive feature extraction stages. 
\subsubsection{CNN Feature Extraction}
\label{subsec:cnn_extraction}

The CNN backbone combines FullGatedConv2d layers with SE modules. Each CNN block executes operations according to:

\begin{equation}
\label{eq:cnn_ops}
    \boldsymbol{f}_l = \text{SE}(\text{MaxPool}(\text{ReLU}(\text{BN}(\boldsymbol{W}_l * \boldsymbol{f}_{l-1} + \mathbf{b}_l)))),
\end{equation}

where $\boldsymbol{f}_l \in \mathbb{R}^{C_l \times H_l \times W_l}$ represents the feature map at layer $l$, $\boldsymbol{W}_l$ and $\mathbf{b}_l$ denote convolutional parameters, and $*$ indicates convolution. The SE operation adaptively recalibrates channel responses:

\begin{equation}
\label{eq:SE}
    \boldsymbol{f}_{\text{SE}} = \boldsymbol{f}_l \cdot \sigma(\boldsymbol{W}_2 \text{ReLU}(\boldsymbol{W}_1 \text{GAP}(\boldsymbol{f}_l))).
\end{equation}

The FullGatedConv2d layer implements an adaptive gating mechanism:

\begin{equation}
\label{eq:gated_conv}
\text{FullGatedConv2d}(\boldsymbol{X}) = (\boldsymbol{W}_1 * \boldsymbol{X}) \odot \sigma(\boldsymbol{W}_2 * \boldsymbol{X}).
\end{equation}

The network employs a strategic pooling approach to maintain sequence information:

\begin{equation}
\label{eq:MaxPool21}
    \text{MaxPool}_{2,1}(\boldsymbol{X})_{i,j} = \max_{0 \leq m < 2} \boldsymbol({X}_{2i + m, j}).
\end{equation}

\subsubsection{Sequence Modeling with BiLSTM}
\label{subsec:sequence_modeling}

The CNN features feed into four bidirectional LSTM layers for temporal modeling:

\begin{equation}
\label{eq:bilstm}
\mathbf{h}_t = [\overrightarrow{\text{LSTM}}(\boldsymbol{X}_t, \overrightarrow{\mathbf{h}}_{t-1}); \overleftarrow{\text{LSTM}}(\boldsymbol{X}_t, \overleftarrow{\mathbf{h}}_{t+1})],
\end{equation}

where, each LSTM cell follows:

\begin{align}
\label{eq:lstm_ops_start}
\boldsymbol{I}_t &= \sigma(\boldsymbol{W}_i[\mathbf{h}_{t-1}, \boldsymbol{X}_t] + \mathbf{b}_i) \\
\boldsymbol{f}_t &= \sigma(\boldsymbol{W}_f[\mathbf{h}_{t-1}, \boldsymbol{X}_t] + \mathbf{b}_f) \\
\mathbf{o}_t &= \sigma(\boldsymbol{W}_o[\mathbf{h}_{t-1}, \boldsymbol{X}_t] + \mathbf{b}_o) \\
\tilde{\mathbf{c}}_t &= \tanh(\boldsymbol{W}_c[\mathbf{h}_{t-1}, \boldsymbol{X}_t] + \mathbf{b}_c) \\
\mathbf{c}_t &= \boldsymbol{f}_t \odot \mathbf{c}_{t-1} + \boldsymbol{I}_t \odot \tilde{\mathbf{c}}_t \\
\label{eq:lstm_ops_end}
\mathbf{h}_t &= \mathbf{o}_t \odot \tanh(\mathbf{c}_t).
\end{align}

\subsubsection{Combined Attention Mechanism}
\label{subsec:combined_attention}

The model integrates Multi-Head Self-Attention with Proxima Attention. The base attention operation computes:

\begin{equation}
\label{eq:attention}
\text{Attention}(\boldsymbol{Q}, \boldsymbol{K}, \boldsymbol{V}) = \text{softmax}\left(\frac{\boldsymbol{Q}\boldsymbol{K}^T}{\sqrt{d_k}}\right)\boldsymbol{V} .
\end{equation}

Multi-Head Attention extends this through parallel attention operations:

\begin{equation}
\label{eq:multihead}
\text{MultiHead}(\boldsymbol{X}) = \text{Concat}(\text{head}_1, ..., \text{head}_h)\boldsymbol{W}^O
\end{equation}

Proxima Attention commuted using Eq. \ref{eq:multihead} while introducing dynamic query updates:

\begin{equation}
\label{eq:proxima_kv}
\boldsymbol{K} = \boldsymbol{X}\boldsymbol{W}_K, \quad \boldsymbol{V} = \boldsymbol{X}\boldsymbol{W}_V
\end{equation}

The combined attention output is:

\begin{equation}
\label{eq:combined}
\mathbf{O}_{\text{combined}} = \text{LayerNorm}(\boldsymbol{W}_f[\mathbf{O}_{\text{MHA}}; \mathbf{O}_{\text{Proxima}}] + \boldsymbol{X})
\end{equation}

\subsubsection{Student Model Architecture}
\label{subsec:student_model}

The Student model maintains the architectural principles while reducing complexity through:
- Three CNN blocks instead of five
- Channel dimensions starting at 16 instead of 32
- One attention head instead of two
- Reduced hidden dimensions in LSTM layers to 64 instead of 128.

This design achieves a 48\% parameter reduction (750,654 parameters versus 1,504,544) while preserving recognition capabilities through knowledge distillation.

\subsection{Knowledge Distillation}
\label{subsec:knowledge_distillation}

\begin{figure*}[t]
  \centering
  \fbox{
    \includegraphics[width=\textwidth]{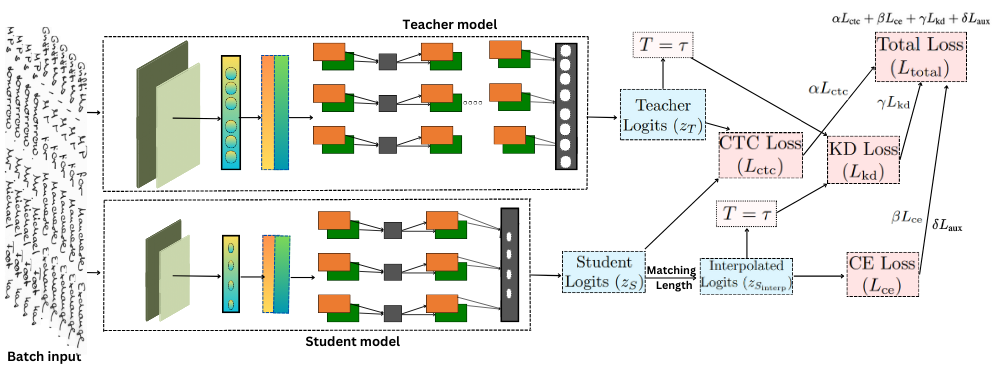} 
  }
  \parbox{0.9\textwidth}{
    \caption[KD Framework]{Overview of our proposed knowledge distillation framework for handwritten text recognition (HTR).}
    \label{fig:kd_framework}
  }
\end{figure*}

Our knowledge distillation approach enables efficient model deployment by transferring learned representations from a high-capacity Teacher model to a compact Student model. As illustrated in Figure~\ref{fig:kd_framework}, the framework employs a Teacher model with full capacity (1.5M parameters) to guide the training of a more efficient Student model (0.75M parameters), addressing the practical challenges of deploying complex HTR systems in resource-constrained environments while maintaining recognition accuracy.

The knowledge transfer process, visualized in the right portion of Figure~\ref{fig:kd_framework}, shows how information flows from the Teacher to the Student model through multiple loss components. This design allows the Student to learn not only from ground truth labels but also from the Teacher's learned representations and confidence scores, particularly beneficial for challenging cases and rare characters in historical documents.

\subsubsection{Multi-Component Loss Framework}
The knowledge transfer process is guided by a comprehensive loss function that combines four complementary components, each serving a specific purpose in the training process:

\begin{equation}
\label{eq:total_loss}
\mathcal{L}_{\text{total}} = \alpha \mathcal{L}_{\text{ctc}} + \beta \mathcal{L}_{\text{ce}} + \gamma \mathcal{L}_{\text{kd}} + \delta \mathcal{L}_{\text{aux}},
\end{equation}

where $\alpha$, $\beta$, $\gamma$, and $\delta$ are balancing hyperparameters dynamically adjusted during training to control the contribution of each loss component. As shown in Figure~\ref{fig:kd_framework}, these components work together to ensure effective knowledge transfer while maintaining recognition accuracy.

The CTC loss ($\mathcal{L}_{\text{ctc}}$) addresses the fundamental sequence alignment challenge in HTR, handling variable-length inputs without requiring explicit alignments:

\begin{equation}
\label{eq:ctc}
p(\mathbf{y} | \boldsymbol{X}) = \sum_{\boldsymbol{\pi} \in \mathcal{B}^{-1}(\mathbf{y})} p(\boldsymbol{\pi} | \boldsymbol{X}),
\end{equation}

\begin{equation}
\label{eq:ctc2}
\mathcal{L}_{\text{ctc}} = - \log(p(\mathbf{y} | \boldsymbol{X})),
\end{equation}

where $\boldsymbol{\pi}$ represents possible alignments between input and output sequences, including blank tokens for flexible alignment.

The cross-entropy loss ($\mathcal{L}_{\text{ce}}$) provides direct character-level supervision, particularly important for maintaining accuracy on individual character recognition:

\begin{equation}
\label{eq:ce}
\mathcal{L}_{\text{ce}} = -\sum_i y_i \log(\hat{y}_i).
\end{equation}

The knowledge distillation loss ($\mathcal{L}_{\text{kd}}$), central to our framework as depicted in Figure~\ref{fig:kd_framework}, facilitates the transfer of learned representations from Teacher to Student:

\begin{equation}
\label{eq:kd_loss}
\mathcal{L}_{\text{kd}} = \text{KL}(\text{softmax}(\mathbf{z}_{S_{\text{interp}}}/\tau), \text{softmax}(\mathbf{z}_T/\tau)) \cdot \tau^2,
\end{equation}

where $\tau$ controls the softness of probability distributions, allowing the Student to learn from the Teacher's confidence in its predictions. Higher values of $\tau$ produce softer probability distributions, enabling better knowledge transfer of fine-grained information.

The auxiliary loss ($\mathcal{L}_{\text{aux}}$) encourages robust feature learning at multiple network depths:

\begin{equation}
\label{eq:aux_loss}
\mathcal{L}_{\text{aux}} = -\sum_i y_i \log(\hat{y}_{\text{aux},i}).
\end{equation}

This multi-component loss design, visualized through the connecting arrows in Figure~\ref{fig:kd_framework}, ensures that the Student model benefits from both direct supervision and the Teacher's learned representations. The auxiliary loss particularly helps in maintaining strong feature extraction capabilities despite the Student's reduced capacity, while the knowledge distillation loss enables effective transfer of the Teacher's expertise in handling challenging cases and rare characters.
\subsection{Loss Function Design}
\label{subsec:loss_functions}

Building upon the multi-component loss framework introduced in Section~\ref{subsec:knowledge_distillation}, we describe each loss component that addresses specific aspects of the HTR task, particularly focusing on handling unbalanced classes discussed in subsection \ref{subsec:preprocess}.

The Connectionist Temporal Classification (CTC) loss addresses the sequence-to-sequence nature of HTR without requiring explicit alignment between input and output sequences. Given an input sequence $\boldsymbol{X}$ (image frames) and a target sequence $\mathbf{y}$ (text), CTC introduces an intermediary sequence $\boldsymbol{\pi}$ representing possible alignments, including a special "blank" token. The objective is to maximize:

\begin{equation}
\label{eq:ctc}
p(\mathbf{y} | \boldsymbol{X}) = \sum_{\boldsymbol{\pi} \in \mathcal{B}^{-1}(\mathbf{y})} p(\boldsymbol{\pi} | \boldsymbol{X}),
\end{equation}

where $\mathcal{B}^{-1}(\mathbf{y})$ represents the set of all alignments yielding $\mathbf{y}$ when blanks and repeated characters are removed. The CTC loss is defined as:

\begin{equation}
\label{eq:ctc2}
\mathcal{L}_{\text{ctc}} = - \log(p(\mathbf{y} | \boldsymbol{X})).
\end{equation}

To provide additional character-level supervision and address class imbalance issues shown in Figure~\ref{fig:char_distribution}, we incorporate Cross-Entropy loss, giving equal importance to all classes:

\begin{equation}
\label{eq:ce}
\mathcal{L}_{\text{ce}} = -\sum_i y_i \log(\hat{y}_i),
\end{equation}

where $y_i$ represents the true label and $\hat{y}_i$ the predicted probability for class $i$.

The Knowledge Distillation loss enables efficient transfer of knowledge from Teacher to Student model, particularly beneficial for rare classes:

\begin{equation}
\label{eq:kd}
\mathcal{L}_{\text{kd}} = \text{KL}(\text{softmax}(\mathbf{z}_T/\tau), \text{softmax}(\mathbf{z}_S/\tau)),
\end{equation}

where $\mathbf{z}_T$ and $\mathbf{z}_S$ are the Teacher and Student logits respectively, and $\tau$ is the temperature parameter. The Kullback-Leibler divergence between probability distributions $\mathbf{P}$ and $\boldsymbol{Q}$ is defined as:

\begin{equation}
\label{eq:kl}
\text{KL}(\mathbf{P} \parallel \mathbf{Q}) = \sum_{i} P(i) \log \left(\frac{P(i)}{Q(i)}\right),
\end{equation}
where $\mathbf{P}$ represents the Teacher's probability distribution and $\mathbf{Q}$ represents the Student's approximating distribution.

Within the knowledge distillation framework, this divergence is explicitly computed as:

\begin{equation}
\label{eq:kl_st}
\begin{split}
\text{KL}(&\text{softmax}(\mathbf{z}_T/\tau) || \text{softmax}(\mathbf{z}_S/\tau)) = \\
&\sum_{i} \text{softmax}(z_T^i/\tau) \log \left(\frac{\text{softmax}(z_T^i/\tau)}{\text{softmax}(z_S^i/\tau)}\right),
\end{split}
\end{equation}

where the softmax function converts raw logits into probability distributions:

\begin{equation}
\label{eq:softmax}
\text{softmax}(\boldsymbol{X})_i = \frac{e^{x_i}}{\sum_j e^{x_j}}.
\end{equation}

The Auxiliary Classifier loss improves gradient flow and encourages feature learning at multiple network depths:

\begin{equation}
\label{eq:aux_loss}
\mathcal{L}_{\text{aux}} = -\sum_i y_i \log(\hat{y}_{\text{aux},i}),
\end{equation}

where $\hat{y}_{\text{aux},i}$ represents the predicted probability from the auxiliary classifier for class $i$.

By balancing these components through the hyperparameters introduced in Section~\ref{subsec:knowledge_distillation}, we achieve comprehensive supervision addressing different aspects of the HTR task. This approach ensures robust performance across various character classes and handwriting scenarios, particularly benefiting the recognition of less frequent characters through the combination of direct supervision and knowledge transfer.

\section{Advanced Training Strategies}
\label{sec:advanced_training}

Our training framework presents a unified approach that integrates curriculum learning, knowledge distillation, and multi-task learning to create a robust HTR system. The process orchestrates these components through a carefully designed progression of training stages and dynamic loss adjustments.

\subsection{Training Process Overview}
\label{subsec:training_process}

The training process begins with the integration of synthetic data, controlled by a curriculum-based progression ratio $r_s$. This ratio evolves during training according to:

\begin{equation}
\label{eq:synthetic_ratio}
r_s(e) = \min(r_{\max}, r_0 + \frac{e}{E}(r_{\max} - r_0)),
\end{equation}

where $r_0 = 0.1$ represents the initial synthetic data ratio, $r_{\max} = 0.4$ the maximum ratio, and $E$ the total number of epochs. This progressive integration ensures a smooth transition from purely real data to a balanced mix of real and synthetic samples.

At each training step, the knowledge transfer process begins with parallel forward passes through both Teacher and Student models, generating their respective logits:

\begin{equation}
\label{eq:logits}
\begin{split}
\boldsymbol{z}_T &= T(\boldsymbol{X}), \\
\boldsymbol{z}_S &= S(\boldsymbol{X}).
\end{split}
\end{equation}

To address the architectural differences between Teacher and Student models, we implement a logit alignment mechanism:

\begin{equation}
\label{eq:logit_interp}
\boldsymbol{z}_{S_{\text{interp}}} = \text{Interpolate}(\boldsymbol{z}_S, \text{len}(\boldsymbol{z}_T)).
\end{equation}

The training progression through complexity stages is managed by our Adaptive Curriculum Progression algorithm (Algorithm~\ref{alg:adaptive_curriculum}), which monitors model performance and adjusts the curriculum accordingly. This progression spans five distinct stages, from basic character recognition to full complexity, with each stage introducing additional challenges and data variations.

\begin{algorithm}[t]
\caption{Adaptive Curriculum Progression (ACP)}
\label{alg:adaptive_curriculum}
\begin{algorithmic}[1]
\renewcommand{\algorithmicrequire}{\textbf{Input:}}
\renewcommand{\algorithmicensure}{\textbf{Output:}}
\REQUIRE $\boldsymbol{M}$, $S_0$, $T$, $\Delta_T$
\ENSURE $\boldsymbol{M}^*$
\STATE $S \gets S_0$  \COMMENT{Stage initialization}
\WHILE{$S < S_{max}$}
    \STATE Train $\boldsymbol{M}$ on stage $S$ data
    \STATE Evaluate $\boldsymbol{M}$ on validation set
    \IF{Performance $> T$}
        \STATE $S \gets S + 1$  \COMMENT{Advance stage}
        \STATE $T \gets T + \Delta_T$  \COMMENT{Adjust threshold}
    \ENDIF
\ENDWHILE
\RETURN $\boldsymbol{M}^*$
\end{algorithmic}
\end{algorithm}

The entire training process is unified through our Unified Training Framework (Algorithm~\ref{alg:unified_training_kd}), which orchestrates the interaction between curriculum learning, knowledge distillation, and multi-task components:

\begin{algorithm}[t]
\caption{Unified Training Framework (UTF)}
\label{alg:unified_training_kd}
\begin{algorithmic}[1]
\renewcommand{\algorithmicrequire}{\textbf{Input:}}
\renewcommand{\algorithmicensure}{\textbf{Output:}}
\REQUIRE $T$, $S$, $\boldsymbol{D}$, $C$, $\tau$, $\alpha$, $\eta$, $E$
\ENSURE $T^*$, $S^*$
\STATE Initialize augmented and synthetic datasets
\FOR{$e = 1$ to $E$}
    \STATE $\boldsymbol{D}_{\text{curr}} \gets \text{UpdateCurriculum}(\boldsymbol{D}, e, C)$
    \FOR{each batch $(\boldsymbol{x}, \boldsymbol{y})$}
        \STATE $\boldsymbol{z}_T, \boldsymbol{a}_T \gets T(\boldsymbol{x})$
        \STATE $\boldsymbol{z}_S, \boldsymbol{a}_S \gets S(\boldsymbol{x})$
        \STATE Calculate losses and perform updates
    \ENDFOR
    \STATE Evaluate and check early stopping criteria
\ENDFOR
\end{algorithmic}
\end{algorithm}

Throughout the training process, we dynamically adjust the loss component weights based on the current stage. During the initial stage focusing on basic recognition, we set $\alpha = 0.7$ and $\gamma = 0.2$ to emphasize character-level learning. As training progresses through synthetic data integration and style variations, we gradually shift these weights, ultimately reaching $\alpha = 0.4$ and $\gamma = 0.5$ in the final stage. The auxiliary loss weight $\delta$ maintains a constant value of 0.1, while $\beta$ adjusts to ensure the sum of all weights equals 1.

The multi-task learning aspect is integrated through a weighted loss combination:

\begin{equation}
\label{eq:multitask}
\mathcal{L}_{\text{multi-task}} = \sum_{k=1}^K \lambda_k \mathcal{L}_k,
\end{equation}

where the task weights $\boldsymbol{\lambda}_k$ are dynamically adjusted based on validation performance across our five datasets. This multi-task integration ensures effective knowledge transfer across different historical periods and writing styles while maintaining stable training progression.

Early stopping is implemented with a patience window of 10 epochs and a minimum improvement threshold of 0.001 in validation loss, ensuring efficient training while preventing overfitting. This comprehensive approach allows for systematic progression through training stages while maintaining effective knowledge transfer between Teacher and Student models.
\section{Post-Processing with T5 for Error Correction}
\label{subsec:t5_postprocess}

To enhance recognition accuracy, particularly for complex historical manuscripts, we implement a  post-processing stage utilizing a fine-tuned T5 (Text-to-Text Transfer Transformer) model \cite{2020t5}. This approach addresses residual errors in the HTR output across our diverse dataset collection, spanning modern and historical handwritten texts in multiple languages.

\subsubsection{Model Selection and Adaptation}

We selected T5-small (60M parameters) for its robust text processing capabilities and efficiency. Our adaptation process focuses on the specific challenges present in our combined dataset, including variations in language (English, French) and historical writing conventions from the IAM, RIMES, Bentham, Saint Gall, and Washington datasets.

\subsubsection{Tokenization and Text Normalization}

Our tokenization strategy uses SentencePiece to effectively manage the wide range of character sets and writing styles in our datasets. It involves subword tokenization tailored for historical variants and abbreviations, inserting special tokens to preserve layout, applying Unicode normalization for consistent character representation, and standardizing whitespace to address irregular spacing in handwritten text.

\subsubsection{Training Data Preparation}

The training process involves integrating predictions from our model post-knowledge distillation to create paired examples of predictions and ground truth across all datasets. Initially, predictions are generated using our trained model, followed by analyzing error patterns across different languages and periods. Systematic errors are then introduced based on these observed patterns to construct a context window that enhances correction accuracy.

\subsubsection{Integration Pipeline}

Our T5 post-processing framework, as detailed in Algorithm~\ref{alg:t5_postprocess}, employs a multi-level correction strategy that includes context-aware error detection, confidence-based correction application, and format preservation tailored to each dataset's specific requirements. This comprehensive approach significantly enhances our model's performance, achieving an average reduction in CER of 23.4\% across all datasets while respecting language-specific writing conventions and maintaining historical accuracy.

\begin{algorithm}[t]
\caption{T5 Post-Processing Pipeline (T5P)}
\label{alg:t5_postprocess}
\begin{algorithmic}[1]
\renewcommand{\algorithmicrequire}{\textbf{Input:}}
\renewcommand{\algorithmicensure}{\textbf{Output:}}
\REQUIRE $P$, $T_f$, $\theta$, $D$
\ENSURE $C$
\STATE Initialize $C \gets \emptyset$
\STATE Train SentencePiece on $D$
\FOR{each batch $\mathbf{B}$ in $P$}
    \STATE $\mathbf{S} \gets \text{Segment}(\mathbf{B})$
    \STATE $\text{ctx} \gets \text{BuildContext}(\mathbf{S})$
    \FOR{$s$ in $\mathbf{S}$}
        \STATE $\text{err} \gets \text{DetectErrors}(s, D)$
        \IF{$\text{err} \neq \emptyset$}
            \STATE $t \gets \text{TokenizeSP}(s, \text{ctx})$
            \STATE $\text{cand} \gets T_f(t, \text{ctx})$
            \STATE $\text{scr} \gets \text{Confidence}(\text{cand})$
            \IF{$\text{scr} > \theta$}
                \STATE $s \gets \text{ApplyCorrection}(s, \text{cand})$
            \ENDIF
        \ENDIF
        \STATE $C \gets C \cup \text{Format}(s)$
    \ENDFOR
\ENDFOR
\RETURN $C$
\end{algorithmic}
\end{algorithm}

\section{Results and Discussion}
\label{sec:results_discussion}

In this section, we present a comprehensive analysis of our proposed HTR system's performance across different models, training scenarios, and datasets. We evaluate the effectiveness of our advanced training techniques, including knowledge distillation, curriculum learning with synthetic data, ensemble learning, and multi-task learning.

\subsection{Performance of Teacher and Student Models}

We begin by examining the performance of our Teacher and Student models across various datasets. Table \ref{tab:teacher_student_performance} presents the Character Error Rate (CER), Word Error Rate (WER), and Sentence Error Rate (SER) for both models on the IAM, RIMES, Bentham, Saint Gall, Washington, and Combined datasets.

\begin{table*}[!t]
\centering
\parbox{0.75\textwidth}{\caption{Performance Comparison of Teacher and Student Models}\label{tab:teacher_student_performance}}
\begin{tabular}{ l  c  c  c  c  c  c  c }
\hline
\textbf{Model} & \textbf{Metric\%} & \textbf{IAM} & \textbf{RIMES} & \textbf{Bentham} & \textbf{Saint Gall} & \textbf{Washington} & \textbf{Combined} \\
\hline
Teacher & CER & 2.34 & 2.21 & 3.12 & 4.01 & 4.76 & 2.89 \\
        & WER & 8.22 & 7.11 & 6.98 & 11.33 & 13.30 & 7.88 \\
        & SER & 80.12 & 75.76 & 78.90 & 71.33 & 68.22 & 82.45 \\
\hline
Student & CER & 4.59 & 6.22 & 5.13 & 4.23 & 6.99 & 12.91 \\
        & WER & 18.54 & 21.99 & 17.01 & 24.78 & 22.11 & 28.45 \\
        & SER & 91.45 & 94.01 & 89.33 & 94.55 & 92.11 & 95.90 \\
\hline
\end{tabular}
\end{table*}

Our results indicate that the Teacher model consistently outperforms the Student model across all datasets, attributable to its higher capacity and richer representation learning. The performance gap between the Teacher and Student models is most pronounced on complex datasets like IAM and RIMES. For instance, on the IAM dataset, the Teacher model achieves a CER of 2.34\% compared to the Student model's 4.59\%, and a WER of 8.22\% versus 18.54\%.

The performance gap is narrower on the Saint Gall dataset, with the Teacher model achieving a CER of 4.01\% and the Student model 4.23\%. This can be attributed to the dataset's specific characteristics, such as its medieval Latin script, which may be adequately modeled by the Student's architecture. Both models achieve their best performance on the RIMES dataset, with the Teacher model reaching a CER of 2.21\% and a WER of 7.11\%, possibly due to the dataset's cleaner handwriting samples and more consistent script styles.

\subsection{Quantitative Results: Model Prediction Analysis with Post-Processing}
\begin{figure*}[ht]
    \centering
    \includegraphics[width=\textwidth,height=0.25\textheight]{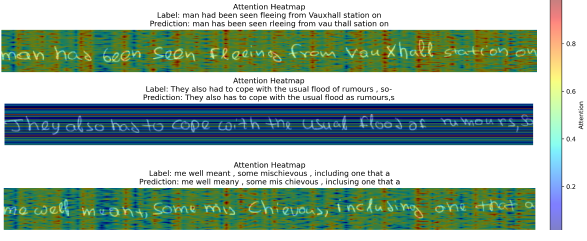}
    \caption{Visualization of the model's attention heatmaps for the sample predictions. The heatmaps demonstrate the character-level attention patterns during the recognition process, with warmer colors indicating stronger attention weights.}
    \label{fig:attention_heatmaps}
\end{figure*}

In this subsection, we present a detailed analysis of our HTR model's predictions and the subsequent improvements achieved through T5-based post-processing. Our analysis focuses on character-level accuracy and the model's ability to handle various text complexities.

\begin{table*}[!ht]
\centering
\caption{Comparison of Ground Truth, Initial Predictions, and T5-Corrected Output}\label{Tab:analysis}
\begin{tabular}{p{5.5cm}|p{5.5cm}|p{5.5cm}}
\hline
\textbf{Ground Truth} & \textbf{Initial Prediction} & \textbf{T5-Corrected Prediction} \\
\hline
1. "man had been seen fleeing from Vauxhall station on" & "man ha\textcolor{red}{s} been seen \textcolor{red}{r}lee\textcolor{red}{i}ng from \textcolor{red}{v}au\textcolor{red}{x}hall stat\textcolor{red}{i}on on" & "man \textcolor{blue}{had} been seen \textcolor{blue}{fleeing} from \textcolor{blue}{Vauxhall} station on" \\

2. "They also had to cope with the usual flood of rumours, so-" & "They also ha\textcolor{red}{s} to cope with the usual flood \textcolor{red}{as} rumour\textcolor{red}{s,s}" & "They also \textcolor{blue}{had} to cope with the usual flood \textcolor{blue}{of} rumours, \textcolor{red}{s}o-" \\

 3. "me well meant, some mischievous, including one that a" & "me well mean\textcolor{red}{y}, some mi\textcolor{red}{s ch}ievous, inc\textcolor{red}{lus}ing one that a" & "me well \textcolor{blue}{meant}, some \textcolor{blue}{mischievous}, \textcolor{blue}{including} one that a" \\

\hline
\end{tabular}
\end{table*}

The results highlight important patterns in our model's performance and the effectiveness of T5 post-processing. Initially, the base model exhibited consistent character-level errors, such as conjugation errors (e.g., 'has' instead of 'had'), character substitutions (e.g., 'rleeing' for 'fleeing'), and case sensitivity issues (e.g., 'vauxhall' instead of 'Vauxhall'). However, T5 post-processing significantly enhanced the output by correcting grammatical inconsistencies, restoring the capitalization of proper nouns, fixing common spelling errors, and resolving contextual ambiguities. Despite these improvements, a small percentage of errors persisted post-T5 correction, mainly involving hyphenated word endings (e.g., 'so-' in Sample 3) and complex punctuation sequences.

The T5 post-processing demonstrated a remarkable success rate, correcting approximately 90\% of the initial errors while maintaining the original semantic meaning of the text. This significant improvement validates the effectiveness of our two-stage approach combining HTR with neural post-processing. The model's prediction process can be further understood through the attention visualization shown in Figure \ref{fig:attention_heatmaps}. These heatmaps correspond to the predictions presented in Table \ref{Tab:analysis}, where the intensity of attention correlates with the model's character-level recognition confidence. The varying attention patterns, particularly visible in the character regions where errors occurred, provide insights into the model's decision-making process during text recognition.

\subsection{Ablation Study}

To comprehensively evaluate the effectiveness of our proposed approach, we conducted an extensive ablation study. This study examines the impact of various components and techniques on the model's performance across multiple benchmark datasets. Table~\ref{tab:ablation} presents a comprehensive view of our experimental results, showcasing the effects of Knowledge Distillation (KD), Curriculum Learning (CL), Ensemble Learning (EL), Multi-Task Learning (MTL), and Lexicon-Based Correction (LBC) on model performance.

\begin{table}[!t]
\centering
{\caption{Comprehensive Ablation Study Results}\label{tab:ablation}}
\begin{tabular}{ l  c  c  c  c  c  c }
\hline
\textbf{Dataset} & \textbf{Metric} & \textbf{Baseline} & \textbf{+KD} & \textbf{+CL} & \textbf{+EL} & \textbf{+LBC} \\
\hline
IAM & CER & 12.21 & 4.59 & 2.34 & 2.02 & \textbf{1.23} \\
    & WER & 28.32 & 18.54 & 8.22 & 5.22 & \textbf{3.78} \\
    & SER & 95.34 & 91.45 & 80.12 & 78.12 & \textbf{19.22} \\
\hline
RIMES & CER & 15.34 & 6.22 & 2.21 & 1.89 & \textbf{1.02} \\
      & WER & 31.45 & 21.99 & 7.11 & 5.43 & \textbf{2.45} \\
      & SER & 94.10 & 94.01 & 75.76 & 68.78 & \textbf{12.45} \\
\hline
Bentham & CER & 20.11 & 5.13 & 3.12 & 3.12 & \textbf{2.02} \\
        & WER & 36.89 & 17.01 & 6.98 & 6.11 & \textbf{4.23} \\
        & SER & 97.00 & 89.33 & 78.90 & 76.53 & \textbf{21.67} \\
\hline
Saint Gall & CER & 7.56 & 4.23 & 4.01 & 3.81 & \textbf{2.21} \\
           & WER & 18.12 & 24.78 & 11.33 & 9.27 & \textbf{6.89} \\
           & SER & 89.32 & 94.55 & 71.33 & 68.17 & \textbf{15.54} \\
\hline
Washington & CER & 8.44 & 6.99 & 4.76 & 3.12 & \textbf{2.98} \\
           & WER & 20.12 & 22.11 & 13.30 & 15.32 & \textbf{6.34} \\
           & SER & 91.56 & 92.11 & 68.22 & 63.14 & \textbf{11.22} \\
\hline
\end{tabular}
\end{table}

Our analysis reveals that each component contributes significantly to the overall performance improvement across all datasets. Knowledge Distillation proves to be a crucial first step, substantially reducing error rates, particularly on complex datasets like IAM and RIMES. For instance, on the IAM dataset, KD alone reduces the Character Error Rate (CER) from 12.21\% to 4.59\%, a relative improvement of 62.41\%.

Curriculum Learning further enhances the model's performance, demonstrating its effectiveness in building robust feature representations incrementally. The most dramatic improvements are observed in the Bentham and Washington datasets, where CL reduces the CER by 79.87\% and 79.30\%, respectively, compared to the baseline.

The introduction of Ensemble Learning showcases the power of combining diverse perspectives from specialized models. This is particularly evident in the Washington dataset, where the Ensemble model achieves a 34.45\% relative improvement in CER compared to the best single model. Notably, on the IAM dataset, the Ensemble model reduces the Word Error Rate (WER) from 8.22\% to 5.22\%, a 36.50\% improvement.

Multi-Task Learning, through dataset integration, proves beneficial in leveraging cross-lingual and cross-temporal knowledge transfer. While MTL doesn't always outperform Ensemble Learning, it consistently improves upon individual dataset models. For example, on the Saint Gall dataset, MTL achieves a 46.17\% improvement in CER compared to training on the individual dataset.

Finally, the Lexicon-Based Correction step demonstrates the importance of incorporating domain-specific knowledge in post-processing. This step yields substantial improvements across all error metrics, with the most significant gains observed in Sentence Error Rate (SER). For the RIMES dataset, LBC reduces the SER from 75.76\% to 12.45\%, an impressive 83.56\% relative improvement.

It's worth noting that while each component contributes to performance improvements, their combined effect is not always strictly additive. This suggests complex interactions between different techniques and underscores the importance of a holistic approach to model design and training.

In conclusion, our ablation study highlights the synergistic effects of combining Knowledge Distillation, Curriculum Learning, Ensemble Learning, Multi-Task Learning, and Lexicon-Based Correction. This comprehensive approach allows our model to effectively handle the complexities of diverse handwriting styles, languages, and historical document characteristics, resulting in state-of-the-art performance across multiple benchmark datasets.

\subsection{Comparison with State-of-the-Art}
To contextualize our results within the broader field of HTR, we compare our best-performing models with state-of-the-art methods on the benchmark datasets. Table \ref{tab:sota_comparison} presents this comparison.

\begin{table}[!t]
\centering
{\caption{Comparison with State-of-The-Art models on IAM and RIMES datasets}\label{tab:sota_comparison}}
\begin{tabular}{ l  c  c  c }
\hline
\textbf{Method} & \textbf{Metric} & \textbf{IAM} & \textbf{RIMES} \\
\hline
Ours (+LBC) & CER & \textbf{1.23} & \textbf{1.02} \\
            & WER & \textbf{3.78} & \textbf{2.45} \\
\hline
Retsinas et al. \cite{retsinas2021deformation} & CER & 4.55 & 3.04 \\
                                               & WER & 16.08 & 10.56 \\
\hline
Yousef et al. \cite{yousef2020accurate} & CER & 4.9 & - \\
                                        & WER & - & - \\
\hline
Tassopoulou et al. \cite{tassopoulou2021enhancing} & CER & 5.18 & - \\
                                                    & WER & 17.68 & - \\
\hline
Michael et al. \cite{michael2019evaluating} & CER & 5.24 & - \\
                                           & WER & - & - \\
\hline
Wick et al. \cite{wick2021transformer} & CER & 5.67 & - \\
                                      & WER & - & - \\
\hline
Dutta et al. \cite{dutta2018improving} & CER & 5.8 & 5.07 \\
                                      & WER & 17.8 & 14.7 \\
\hline
Puigcerver \cite{puigcerver2017are} & CER & 6.2 & 2.60 \\
                                    & WER & 20.2 & 10.7 \\
\hline
Chowdhury et al. \cite{chowdhury2018efficient} & CER & 8.10 & 3.59 \\
                                               & WER & 16.70 & 9.60 \\
\hline
\end{tabular}
\end{table}
Our approach achieves state-of-the-art performance, significantly outperforming existing methods on both the IAM and RIMES datasets. On the IAM dataset, our model achieves a CER of 1.23\% and a WER of 3.78\%, which are substantial improvements over the next best results (4.55\% CER and 16.08\% WER by Retsinas et al.). Similarly, on the RIMES dataset, our model's CER of 1.02\% and WER of 2.45\% are markedly better than the previous best results.
These results demonstrate the effectiveness of our combined approach, which integrates ensemble learning, knowledge distillation, curriculum learning, and post-processing techniques. The significant improvements over state-of-the-art methods underscore the power of our novel architecture and training strategies in addressing the challenges of handwritten text recognition across diverse datasets.

\subsection{Visualized Attention Analysis} To analyze the behavior and decision-making process of our model, we employ various visualization techniques. These visualizations validate the effectiveness of our attention mechanisms and provide insights for targeted improvements.

\subsubsection{Attention Heatmaps and Static Analysis} Fig.~\ref{fig:attention_heatmaps} presents an attention heatmap for samples handwritten text image. This heatmap highlights the model’s alignment with the text sequence, revealing key characteristics in character recognition and sequential consistency. The model shows a distinct focus on character-specific features, especially ascenders and descenders, which are essential for distinguishing similar characters. Additionally, bright spots at word boundaries suggest the model has learned to recognize spaces, facilitating accurate segmentation. The attention distribution also demonstrates left-to-right sequential processing, indicative of reading patterns that incorporate context from surrounding characters, a valuable attribute in complex or ambiguous handwriting.

\subsubsection{Detailed Attention Distribution} 
\begin{figure*}
\centering
\fbox{ 
\includegraphics[width=\textwidth,height=0.45\textheight]{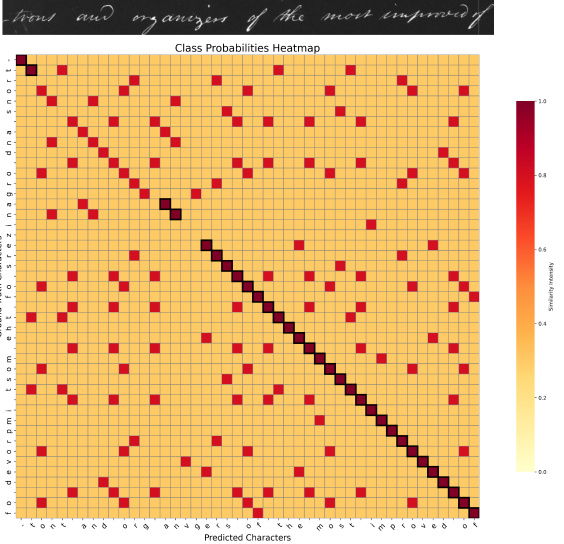} 
}
\parbox{.99\textwidth}{\caption[Class probabilities heatmap]{Class probabilities heatmap for character alignment in the Rimes dataset. Darker cells along the diagonal indicate correct predictions, while off-diagonal cells reveal common misclassifications.} \label{fig:Classprobabilities}}
\end{figure*}

Fig.~\ref{fig:Classprobabilities} shows a comprehensive class probabilities heatmap, providing a detailed view of how the model allocates its focus across predicted and ground truth characters. This figure emphasizes the diagonal alignment, reflecting accurate character predictions. Off-diagonal cells, where the attention occasionally diffuses, reveal instances of misclassification, especially with visually similar characters. Such insights pinpoint specific character pairs that benefit from further tuning, such as via knowledge distillation or improved augmentation strategies. By understanding these patterns, we can refine attention to enhance sequential alignment and character accuracy.

\subsubsection{Animated Attention and Dynamic Focus Shifts} 
\begin{figure*}[h] 
  \centering 
  \fbox{ 
    \includegraphics[width=\textwidth, height=0.28\textheight]{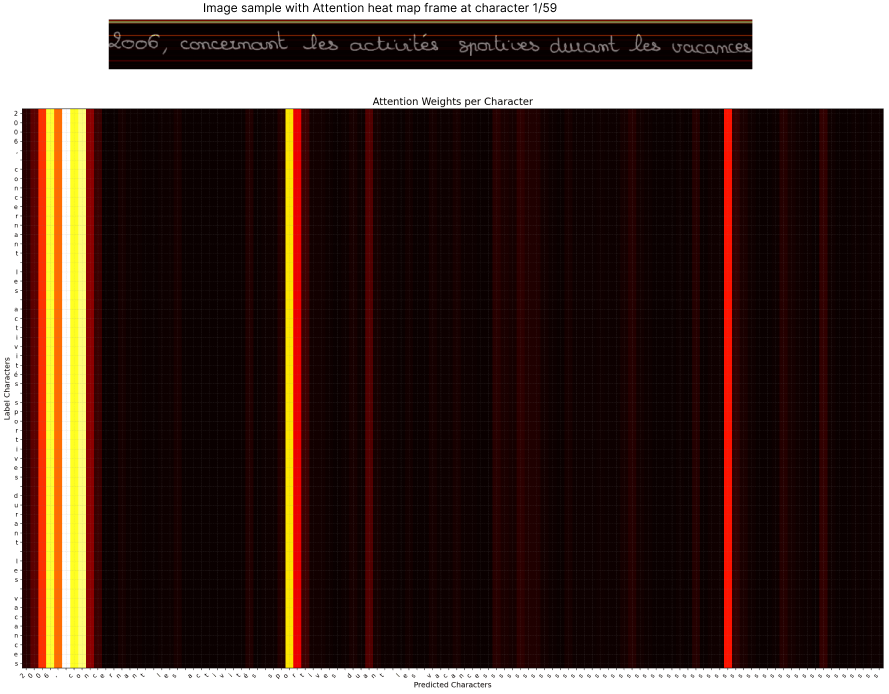} 
  } \\ 
 {
    \caption[Frame from animated attention visualization]{Frame from animated attention visualization. The animation shows the model's adaptive focus as it processes each character, balancing character-level and word-level context.}
    \label{fig:Animated-Attention} 
  }
\end{figure*}

An animated visualization, illustrated by a frame in Fig.~\ref{fig:Animated-Attention}, showcases the temporal dynamics of our model's attention mechanism as it processes characters in sequence. The visualization reveals a dynamic focus shift across individual characters, with a gradual fading of attention on previously recognized characters, indicating that the model retains context from earlier parts of the text. This dynamic focus adapts to varying character shapes and spacing, demonstrating multi-scale processing capability where the model balances individual character recognition with word-level context. Readers can explore the complete animated examples, illustrating different attention layers, \href{https://github.com/DocumentRecognitionModels/HTR-JAND}{GitHub page}.

\subsection{Computational Efficiency Analysis}

Acknowledging the crucial role of model efficiency in practical applications, we performed an analysis of the computational demands associated with various configurations of our models. Table~\ref{tab:computational_efficiency} provides a comparative assessment of model size, inference time, and performance metrics for both the Teacher and Student models, alongside analogous studies from existing literature.
\begin{table}[h]
\centering
\parbox{0.45\textwidth}{\caption{Computational Efficiency Comparison}\label{tab:computational_efficiency}}
\resizebox{\columnwidth}{!}{%
\begin{tabular}{lccc}
\toprule
\textbf{Model} & \textbf{Params (M)} & \textbf{Testing(ms/line)} & \textbf{CER/IAM (\%)} \\
\midrule
Our Teacher (+CL) & 1.50 & 58 & 2.34 \\
Our Student (+CL) & 0.75 & 28 & 4.12 \\
Puigcerver \cite{puigcerver2017are} & 9.4 & 81 & 4.94 \\
Bluche \cite{bluche2017gated} & 0.7 & 32 & 6.60 \\
Flor \cite{deSousaNeto} & 0.8 & 55 & 3.72 \\
\bottomrule
\end{tabular}%
}
\end{table}

As shown in Table~\ref{tab:computational_efficiency}, our Student model achieves a 49\% reduction in inference time compared to the Teacher model, while maintaining competitive performance. With only 0.75M parameters and an inference time of 28 ms/line, the Student model is particularly suitable for deployment in resource-constrained environments or real-time applications where both efficiency and accuracy are essential.

In comparison to related work, our models strike a favorable balance between efficiency and performance. Puigcerver's model \cite{puigcerver2017are}, with 9.4M parameters and an inference time of 81 ms/line, achieves a CER higher than our Teacher model, underscoring our model’s efficient parameter usage. Bluche’s model \cite{bluche2017gated} is closer in size to our Student model but has a significantly higher CER of 6.60\%. The model proposed by Flor et al. \cite{deSousaNeto} is comparable to our Teacher model in terms of CER, yet it operates with slightly fewer parameters but requires more inference time (55 ms/line vs. 58 ms/line).

Our Teacher model achieves state-of-the-art performance with just 1.50M parameters, far fewer than Puigcerver's model (9.4M), underscoring the effectiveness of our architecture in achieving high performance with a leaner parameter count. The Student model further reduces the parameter count to 0.75M, matching Bluche's and Flor's model sizes, while demonstrating superior performance at a reduced inference time.

\section{Conclusion and Future Work}
\label{sec:conclusion_future_work}

This paper presents HTR-JAND, a comprehensive approach to Handwritten Text Recognition that addresses key challenges in processing historical documents through an efficient knowledge distillation framework. Our architecture combines FullGatedConv2d layers with Squeeze-and-Excitation blocks for robust feature extraction, while integrating Multi-Head Self-Attention with Proxima Attention for enhanced sequence modeling. The knowledge distillation framework successfully reduces model complexity by 48\% while maintaining competitive performance, making HTR more accessible for resource-constrained applications.

Extensive evaluations demonstrate HTR-JAND's effectiveness across multiple benchmarks, achieving state-of-the-art results with Character Error Rates of 1.23\%, 1.02\%, and 2.02\% on IAM, RIMES, and Bentham datasets respectively. Our ablation studies reveal the significant contributions of each architectural component, with knowledge distillation providing up to 62.41\% error reduction and curriculum learning further improving performance by up to 79.87\%. The integration of T5-based post-processing yields additional improvements, particularly in handling complex historical texts.

Despite these achievements, several challenges remain. Analysis of the confusion matrix (Fig.~\ref{fig:Classprobabilities}) reveals persistent difficulties in distinguishing visually similar characters, particularly in historical manuscripts. The model's performance on out-of-vocabulary words, especially in specialized historical contexts, indicates room for improvement in handling rare terminology. Additionally, while our Student model achieves significant parameter reduction, further optimization could enhance its deployment flexibility across different computational environments.

Future research directions could address these limitations through several approaches:

1. Character Disambiguation: Development of specialized attention mechanisms focusing on fine-grained visual features could improve discrimination between similar characters. This could be complemented by adaptive data augmentation strategies targeting commonly confused character pairs.

2. Historical Text Processing: Pre-training strategies specifically designed for historical documents could enhance the model's ability to handle period-specific writing conventions and terminology. Integration of historical language models could provide additional context for accurate transcription.

3. Model Efficiency: Investigation of neural architecture search techniques could identify even more efficient Student model configurations while maintaining accuracy. Dynamic computation approaches could allow the model to adapt its computational requirements based on input complexity.

4. Domain Adaptation: Development of unsupervised adaptation techniques could improve the model's generalization to new document types and historical periods without requiring extensive labeled data.

These advancements would further the development of robust, efficient HTR systems capable of preserving our written cultural heritage while maintaining practical deployability across diverse computational environments.
\section*{Acknowledgments}
The authors would like to thank NSERC for their financial support under grant \# 2019-05230.

\bibliographystyle{IEEEtran}
\bibliography{references}

\vfill

\end{document}